\documentclass[conference]{IEEEtran}
\IEEEoverridecommandlockouts
\usepackage{cite}
\usepackage{amsmath,amssymb,amsfonts}
\usepackage{algorithmic}
\usepackage{graphicx}
\usepackage{textcomp}
\usepackage{xcolor}

\usepackage{subcaption}
\usepackage{amsmath}
\usepackage{multirow}
\def\BibTeX{{\rm B\kern-.05em{\sc i\kern-.025em b}\kern-.08em
    T\kern-.1667em\lower.7ex\hbox{E}\kern-.125emX}}
\begin{document}

\title{An Interpretable Deep Semantic Segmentation Method for Earth Observation\\
\thanks{The project is co-funded by the ESA (European Space Agency).}
}

\author{\IEEEauthorblockN{1\textsuperscript{st} Ziyang Zhang}
\IEEEauthorblockA{\textit{School of Computing and Communications} \\
\textit{Lancaster University}\\
Lancaster, UK \\
z.zhang51@lancaster.ac.uk}
\and
\IEEEauthorblockN{2\textsuperscript{nd} Plamen Angelov}
\IEEEauthorblockA{\textit{School of Computing and Communications} \\
\textit{Lancaster University}\\
Lancaster, UK \\
p.angelov@lancaster.ac.uk}
\and
\IEEEauthorblockN{3\textsuperscript{rd} Eduardo Soares}
\IEEEauthorblockA{\textit{School of Computing and Communications} \\
\textit{Lancaster University}\\
Lancaster, UK \\
e.almeidasoares@lancaster.ac.uk}
\and
\IEEEauthorblockN{4\textsuperscript{rd} Nicolas Longepe}
\IEEEauthorblockA{\textit{Phi-Lab Explore Office} \\
\textit{European Space Agency}\\
Frascati, Italy \\
nicolas.longepe@esa.int}
\and
\IEEEauthorblockN{5\textsuperscript{rd} Pierre Philippe Mathieu}
\IEEEauthorblockA{\textit{Phi-Lab Explore Office} \\
\textit{European Space Agency}\\
Frascati, Italy \\
pierre.philippe.mathieu@esa.int}

}

\IEEEpubid{\begin{minipage}{\textwidth}\ \\[12pt] 978-1-6654-5656-2/22/\$31.00 \copyright 2022 \end{minipage}}

\maketitle

\begin{abstract}
Earth observation is fundamental for a range of human activities including flood response as it offers vital information to decision makers. Semantic segmentation plays a key role in mapping the raw hyper-spectral data coming from the satellites into a human understandable form assigning class labels to each pixel. Traditionally, water index based methods have been used for detecting water pixels. More recently, deep learning techniques such as U-Net started to gain attention offering significantly higher accuracy. However, the latter are hard to interpret by humans and use dozens of millions of abstract parameters that are not directly related to the physical nature of the problem being modelled. They are also labelled data and computational power hungry. At the same time, data transmission capability on small nanosatellites is limited in terms of power and bandwidth yet constellations of such small, nanosatellites are preferable, because they reduce the revisit time in disaster areas from days to hours. Therefore, being able to achieve as highly accurate models as deep learning (e.g. U-Net) or even more, to surpass them in terms of accuracy, but without the need to rely on huge amounts of labelled training data, computational power, abstract coefficients offers potentially game-changing capabilities for EO (Earth observation) and flood detection, in particular. In this paper, we introduce a prototype-based interpretable deep semantic segmentation (IDSS) method, which is highly accurate as well as interpretable. Its parameters are in orders of magnitude less than the number of parameters used by deep networks such as U-Net and are clearly interpretable by humans. The proposed here IDSS offers a transparent structure that allows users to inspect and audit the algorithm's decision. Results have demonstrated that IDSS could surpass other algorithms, including U-Net, in terms of  IoU (Intersection over Union) total water and Recall total water. We used WorldFloods data set for our experiments and plan to use the semantic segmentation results combined with masks for permanent water to detect flood events. 
\end{abstract}


\section{Introduction}

Flood is one of the most catastrophic weather events as it caused about 7 million fatalities in the twentieth century \cite{merz2021causes}. The average annual loss generated by floods are estimated at over US$\$$ 100 billion (2015) \cite{wannous2017united}. Recent findings claim that the exposure to floods is expected to grow three times by 2050 due to increases in population and economic assets in flood-prone areas \cite{jongman2012global}. Depending on the socio-economic scenario, human losses from flooding are projected to rise by 70–83$\%$ and direct flood damage by 160–240$\%$ relative to 1976–2005 \cite{dottori2018increased}. Detecting flooding and its associated impacts is critical to effective risk reduction \cite{burningham2008ll, willner2018global}.

The advent of Copernicus of ESA (European Space Agency) and the launch of several Sentinels satellites have provided massive amounts of data for a range of Earth observation missions, including flood detection \cite{onoda2017satellite}. SAR (Synthetic Aperture Radar) images and optical remote sensing images are often used for flood mapping, but the optical remote sensing images is preferable due to its mature processing and analysis techniques, while SAR images suffer from noise and information loss problems \cite{zhang2021unsupervised}. The Sentinel-2 satellite constellation is often used for flood detection due to the availability of Multi-Spectral Imagers, shorter revisit times (5 days), and higher spatial resolution (10m for some bands) \cite{cavallo2021continuous}.  

Flood mapping traditionally relies on water indices, such as NDWI \cite{mcfeeters1996use} and MNDWI \cite{xu2006modification}. However, these methods are threshold-based, which requires expert knowledge. In recent years, machine learning algorithms have also been applied to flood detection. \cite{cordeiro2021automatic}  uses clustering algorithm and a combination of several water indexes to identify the water areas. However, the coastal water, clouds, shadows and snow pixels had to be removed in advance, which greatly increases the workload. 

The recent work from \cite{mateo2021towards} investigates how a constellation of small nano satellites assembled from commercial off-the-shelf (COTS) hardware, also known as CubeSats, could be used for disaster response in the case of floods. The authors proposed the use of deep learning algorithms to produce multi-class segmentation with high accuracy on-board of a very cheap satellite hardware. Although the deep learning approaches proposed for this challenging task have produced great results in terms of accuracy, they are black-box and are extremely difficult to audit \cite{angelov2021explainable}. 

In this paper, a new prototype-based approach is proposed called interpretable deep semantic segmentation (IDSS), which extends the recently introduced explainable Deep Neural Network (xDNN) \cite{angelov2020towards} through a new clustering and decision mechanism. The prototype-based nature of IDSS allows clear interpretability of its decision mechanism  \cite{bien2011prototype}. This is important for the human-in-the-loop process involved in this application. Results demonstrate that IDSS is able to surpass xDNN and U-Net in terms of IoU and recall for water detection.

\section{Literature review}
\subsection{Water/Flood mapping}

Water/Flood mapping requires semantic segmentation for allocating a class label to each multidimensional pixel. Traditionally, different water index-based methods are being used to determine liquid water. They represent ratios of different spectral bands from the raw satellite signal that can characterize the water absorption. The rationale behind the usefulness of the water indices is that the water absorbs energy in the near infrared (NIR) and short-wave infrared (SWIR) wavelengths, making it significantly different from other objects on the ground \cite{ji2009analysis}. The most widely used water index is NDWI \cite{mcfeeters1996use}. Other water indices also exists such as MNDWI \cite{xu2006modification}, WNDWI \cite{ guo2017weighted} and AWEI \cite{feyisa2014automated}. However, such methods often require manual setting of thresholds, which is challenging and the choice does influence the result significantly.

Machine learning methods were also applied for flood mapping. For example, K-means was used to perform clustering based on Synthetic Aperture Radar (SAR) optical features and thresholds were applied to further perform classification by \cite{ landuyt2020flood}. The SVM classifier was used as a reference label since it is considered to provide better results than the thresholding methods \cite{ sekertekin2021survey}. KNN was used to perform per-pixel classification based on the water index by \cite{ pan2020comparative}.

More recently, with the rapid development of deep learning, convolutional neural networks started to be used more widely for Earth Observation and flood mapping, in particular. For example, the fully convolutional neural network Resnet50 trained on Sentinel-1 satellite images was used to segment permanent water and flood by \cite{ bonafilia2020sen1floods11}, U-Net and a simple CNN were trained on Sentinel-2 satellite images used to perform the onboard flood segmentation task by \cite{ mateo2021towards}. The main advanatge of using deep convolutional networks is their high levels of accuracy measured primarily by IoU (intersection over union) and recall characteristics for this particular problem. They also offer powerful latent features extraction capability. However, the downside is that they require large amount of labeled training data and computational power and most of all, they lack interpretability. They are often considered as "black box" models because they have millions of abstract model parameters (weights) which have no direct physical meaning and are hard to interpret or check if affected by noise or adversarial actions especially when transmitted. Attempts have been made to provide some explainability, but these are mostly \textit{post-hoc} partial solutions or surrogate models. Therefore, currently, there is a powerful trend aiming to develop explainable or interpretable-by-design alternatives that are as powerful as such Deep Neural Networks, yet offer human-intelligible models \cite{rudin2019stop}, \cite{angelov2020towards}, \cite{angelov2021explainable}. 

The method proposed in this paper offers visual and linguistic forms of interpretation based on prototypes, offering layers with clear interpretation as well as a linguistic \textit{IF... THEN } rules and clear decision making process based on similarity. These forms of representation of the model can be inspected by a human and have clear meaning.
The prototypes can also be visualized using RGB colours as well as raw features such as NIR, SWIR, etc. which are easy to interpret by a human. 

In this next sub-section, prototype-based machine learning methods are briefly reviewed because they are underestimated and often overlooked, but they do offer clear advantages in terms of interpretability and high levels of accuracy and performance.

\subsection{Prototype-based models}

Prototype-based machine learning methods have similar concepts as some methods from cognitive psychology and neuroscience in regards to comparisons of new observations or stimuli to a set of prototypes \cite{biehl2016prototype}. Prototype-based machine learning models have been attracting much attention due to their easy to understand decision making processes and interpretability. K nearest neighbor \cite{fix1989discriminatory} and 
K-means \cite{ hartigan1979algorithm} are the most representative algorithms based on prototypes. The most typical prototype-based neural network algorithm is the Radial-basis Function (RBF) method \cite{buhmann2000radial} which is like a bridge between neural networks and linguistic \textit{IF...THEN} rules. There are also other prototype-based machine learning algorithms such as fuzzy c-means clustering \cite{bezdek1984fcm} and Learning Vector Quantization (LVQ) \cite{ kohonen1995learning}. 

Recently, with the rapid development of deep learning algorithms, prototype-based models have a tendency to integrate with neural networks. Such as ProtoPNet \cite{chen2019looks}, xDNN \cite{angelov2020towards} and a nonparametric segmentation framework \cite{ zhou2022rethinking}. The IDSS method proposed in this paper differs from these works because xDNN \cite{angelov2020towards} does not perform clustering to generate prototypes and has not been applied to flood mapping while the work by \cite{ zhou2022rethinking} only look for prototypes in the embedding space, which greatly reduces the interpretability of the model. The method proposed in this paper not only outperforms the state-of-the-art deep convolutional networks such as UNet and SCNN as well as various water-indices-based models, but also, greatly improves the interpretability of the model by further finding the mean value in the raw features space corresponding to the prototypes (cluster centers) in the latent features space and can be described by linguistic \textit{IF... THEN } rules.

\begin{figure*}[htb]
\centerline{\includegraphics[width=0.7\textwidth]{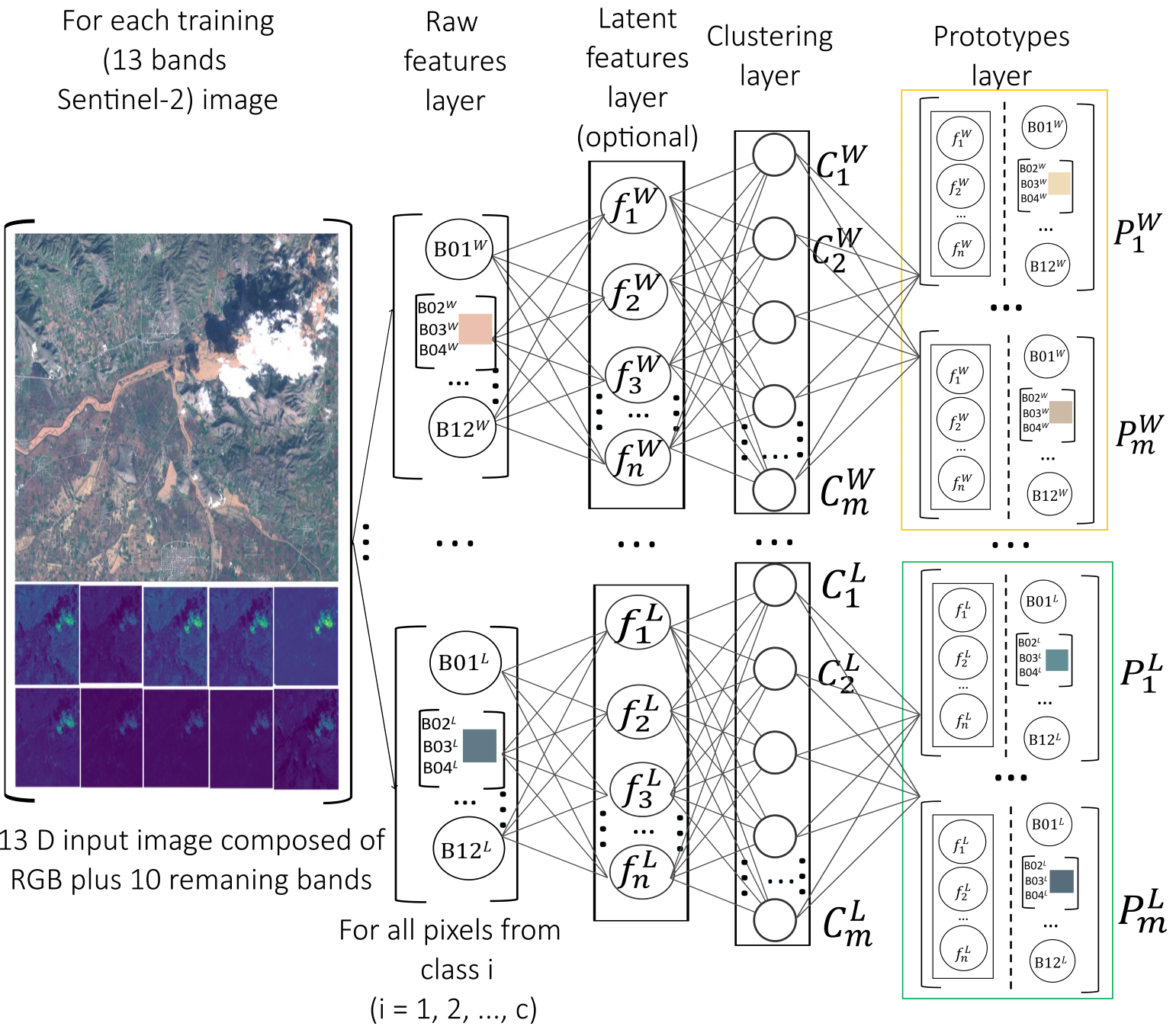}}
\caption{Training process of IDSS}
\label{TrainingF}
\end{figure*}

\section{Proposed Method}
In this paper a new method is proposed named interpretable deep semantic segmentation (IDSS). It is prototype-based unlike the majority of the alternatives which are highly parameterized. IDSS considers that each class can be divided into different sub-classes in the latent feature space. These can result from clustering the data of a given class (e.g. water) or by selecting peaks of the data distribution as in the xDNN approach \cite{angelov2020towards}. Clustering can be done in the raw or latent features space. IDSS is using clustering. In this paper the mini-batch K-means \cite{sculley2010web} clustering method is used on the latent features space without loss of generality. The correspondence between the latent feature space representation and the raw feature space representation for the cluster centers is being preserved by performing the mini-batch K-means algorithm on the latent feature space, but performing all averaging (mean) operations to both, the raw and the latent feature space representations. In this way, all the decisions about reallocations of points to clusters are made based on the latent feature representation; however, each center is being exactly represented in both, the raw and latent feature space. The raw feature space representation is more suitable to produce an interpretable linguistic \textit{IF...THEN} form of the prototypes/means. This semantic form of representation facilitates the human perception and analysis of the decisions made by the algorithm.      

The training architecture of the proposed IDSS method is shown in Fig. \ref{TrainingF}. It is described in more detail below:

\begin{figure*}[htb]
\centerline{\includegraphics[width=0.7\textwidth]{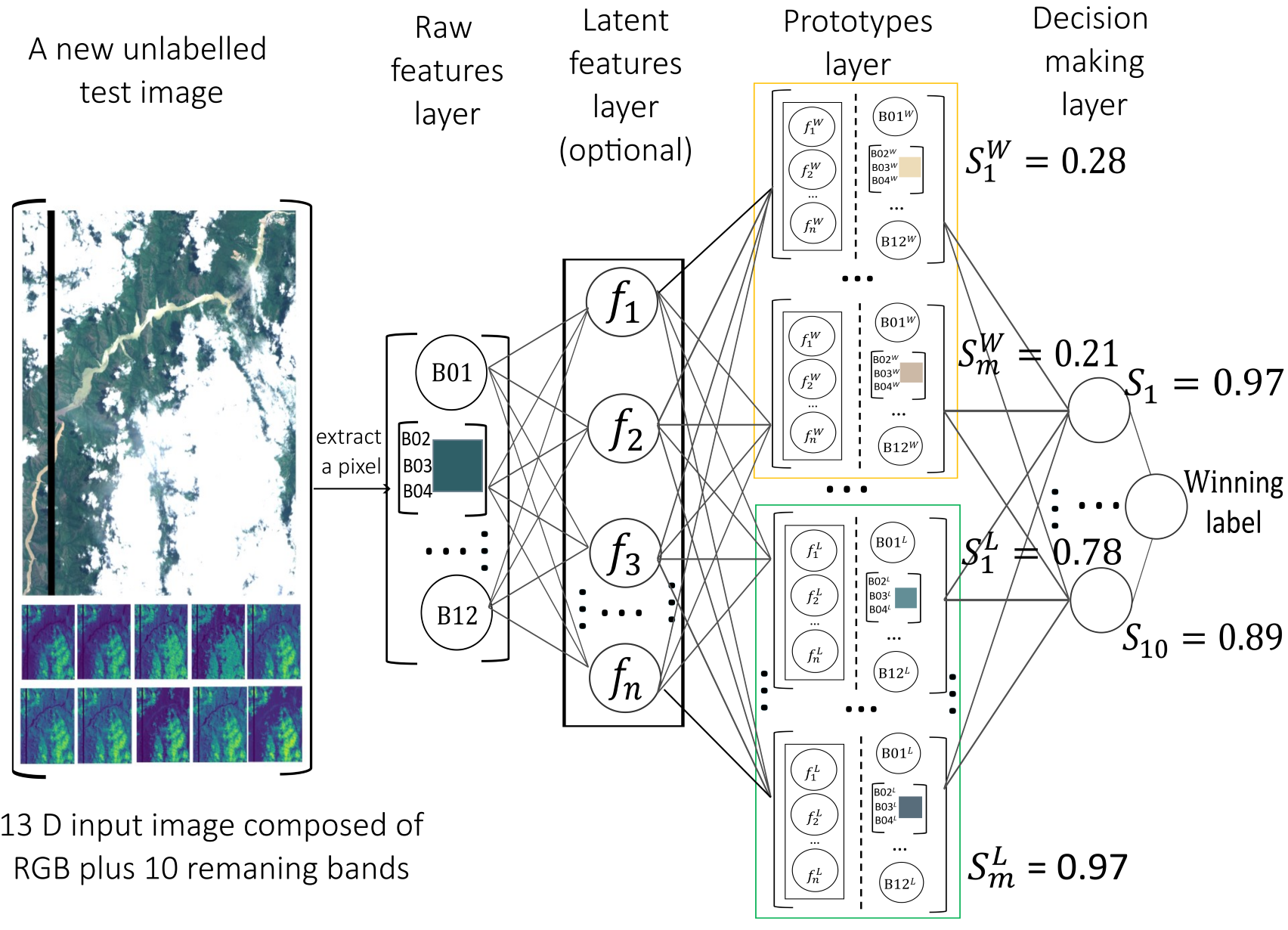}}
\caption{Validation process of IDSS}
\label{ValidationF}
\end{figure*}

1)	\textbf{Raw features layer}: The Raw features layer is responsible for extracting each pixel, $x$ from the image and assigning it to different classes using corresponding labels since our algorithm is trained per class. In this layer the raw features (e.g. the 13 features from the hyper-spectral signal of the Sentinel-2 satellite images as described in \textit{\textbf{Table \ref{Table1}}}) are extracted per pixel. This vector of raw features is further passed to the next later which may be clustering or (optionally) - feature transformation.

\begin{table}[htb]
\caption{Hyper-spectral 13-band signal from the Sentinel-2 satellite (vector of raw features)}
\begin{center}
\begin{tabular}{|c|c|c|}
\hline
Name & Description         & Resolution (m)  \\ 
\hline
B01  & Coastal aerosol     & 60                         \\
B02  & Blue                & 10                         \\
B03  & Green               & 10                    \\
B04  & Red                 & 10                        \\
B05  & Vegetation red edge & 20                      \\
B06  & Vegetation red edge & 20                        \\
B07  & Vegetation red edge & 20                        \\
B08  & NIR                 & 10                        \\
B8A  & Narrow NIR          & 20                       \\
B09  & Water vapour        & 60                        \\
B10  & SWIR                & 60                         \\
B11  & SWIR                & 20                        \\
B12  & SWIR                & 20                        \\ 
\hline
\end{tabular}
\end{center}\label{Table1}
\end{table}

2)	\textbf{Latent features layer (optional)}: This layer is optional. For a better performance, the raw features can be transformed, e.g. by orthogonalization and optimization using PCA \cite{jolliffe2016principal}. Alternatively, deep feature representations using pre-trained convolutional neural network such as Segnet \cite {badrinarayanan2017segnet} or U-Net \cite {ronneberger2015u} can be used. It is also possible to directly feed raw features into the next clustering layer. In this paper, U-Net Deep Neural Network that is pre-trained on Worldfloods dataset \cite{mateo2021towards} is used to perform feature extraction. The input to the U-Net is formed by the raw 13 bands hyper-spectral Sentinel-2 pixel values from the training images and the penultimate layer of the U-Net is used for feature extraction. The output is, thus, a 64-dimensional feature vector. 

For all pixels, $x_j^i$ (\textit{j = 1, 2,..., N}; where \textit{N} denotes the number of pixels of a given image) for each class $i$ (where \textit{i = 1, 2, ..., c}), the U-Net is used to map the raw features vector $x_j^i$ into a \textit{n}-dimensional latent feature space $\phi$, where each feature is represented as $F_j^i$. In this paper, $F_j^i \in R ^{64}$, i.e. $n=64$, since the output of the penultimate layer of the U-Net is a 64-dimensional vector. Then, L2 normalization is performed to the (latent) features as shown below:

\begin{equation}
    F_j^i = \dfrac{F_j^i}{||F_j^i||}
\end{equation}

For \textit{i = 1, 2, ..., c}, \textit{j = 1, 2,..., N}.

3)	\textbf{Clustering layer}: The Clustering layer is responsible for two main tasks: applying clustering to the latent features and finding the cluster centers, $C_m^i$ where 
\textit{m } represents the number of cluster centers per class. Each cluster center does have a corresponding representation in the raw features ($x_j^i$) domain which can serve as a prototype providing a clear interpretability. 

The mini-batch K-means \cite{sculley2010web} clustering method is used in this paper without lose of generality. This particular version of the K-means approach is reported to be more suitable for handling large data sets and can be updated online. The main output of the K-means algorithm is the set of cluster centers, $C_m^i$ which minimize the distance between clusters in terms of the latent features used, $F_j^i \in F$.

The choice of the number of centers m is important for the performance of the mini-batch K-means algorithm. In this paper, m is set to 500, which means that each of the land, water and clouds classes have 500 corresponding prototypes.

\begin{figure}[htbp]
\centerline{\includegraphics[width=0.5\textwidth]{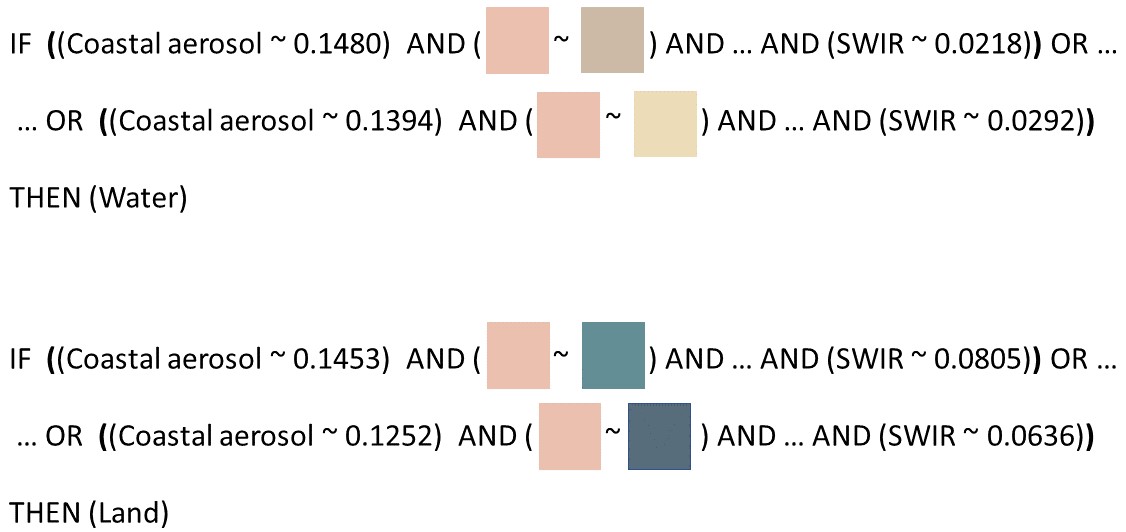}}
\caption{Linguistic rules of IDSS}
\label{Linguistic}
\end{figure}

4)	\textbf{Prototypes layer}: The prototypes are considered to correspond to the identified centres, $C_m^i$; however, they are being expressed in terms of raw features. Since the use of latent features is optional (in particular, one of the results from this study shown in Table \ref{Table3} is using raw features and no latent ones), prototypes may directly correspond to the centers. In cases, when latent features are being used, a transformation is needed which is a result of finding the means at each step of the k-means algorithm in terms of both, raw and latent features. It has to be stressed, however, that the decision how to reallocate data samples between clusters which is a key part of the K-means approach is based on latent features (if they are being used). The raw features are convenient to communicate the result to the human users. In particular, the prototypes can generate linguistic \textit{IF ... THEN} rules, as shown in Fig. \ref{Linguistic}, where $\sim$ denotes "is similar to" and "Coastal", "SWIR", etc., represent the features (e.g. the 13 bands of the Sentinel-2 signal) for a single pixel. Such linguistic rules are easy to understand, explain and use by humans because they are intuitive. They can be applied to unlabeled test images as a tool to understand the decision making process for deciding which class a particular pixel is more likely to belong to. One such linguistic rule can be formed per prototype connecting individual terms such as (SWIR~...), (Coastal ~...) etc. with logical conjunction (AND). In particular, each term concerns an individual raw feature (SWIR, Coastal,...). Brought together with the logical conjunction, AND operator they form an \textit{IF...THEN} rule where the THEN condition defines the class label (e.g. Water). In this paper, 500 clusters are considered per class and, therefore, 500 such \textit{IF...THEN} rules are being generated per class. All these 500 rules can be combined together using logical disjunction, OR operator resulting in a single (but large) linguistic rule per class. Although such rule is large it is quite clear and easy to interpret, understand and explain to a human. In a summary, the outcome can be eiter one large linguistic rule per class or multiple smaller rules per class.

The training  procedure of IDSS is summarised bellow:

The validation architecture of IDSS is responsible for making predictions on new unlabeled Sentinel-2 test images, as shown in  Fig. \ref{ValidationF}, described in detail below:

1) \textbf{Raw features layer}: This layer is responsible for extracting each pixel \textit{x} from the unlabelled Sentinel-2 test image quite similarly to the similar layer of the Training architecture.

2) \textbf{Latent features layer (optional)}: This layer is the same as the training architecture, 64-dimensional feature value is extracted from the penultimate layer of the U-Net. then, L2 normalization is performed same as in the training architecture.

3) \textbf{Prototypes layer}: This layer is responsible for calculating the similarity to each prototype from each class. Similarity, \textit{S} is calculated in the latent $F_j$ (or raw, $x_j$) feature space using exponential kernel:

\begin{equation}
    S(F^*,P_j^i) = exp(-||F^*-P_j^i||^2)
\end{equation}

where $F^*$ denotes a pixel from the test (unlabeled, new) image represented in the form of the latent feature and \textit{ P }denotes the $j^{th}$ prototypes of the $i^{th}$ class.

4)	\textbf{Decision-making layer}: The decision-making layer is responsible for assigning labels to unknown pixels. In this study, K nearest neighbours, also called the "few winners take all" method, is used to make the final decision. In this study, K is set to 10, which means that the labels corresponding to the ten prototypes with the highest similarity to the unlabeled pixels are used to perform decision-making. The label can be obtained by the formula (3).

\begin{equation}
    L(F^*) = \mathop{\arg\max}\limits_{\beta \in c}\sum_{i=1}^{K}\delta(\beta,l(F_i))
\end{equation}

where \textit{L} denotes the winning label, \textit{$l(F_i)$} denotes the label of k nearest neighbors of $F^*$, $\delta(\beta, l(F_i))  = 1 $ if $\beta = l(F_i)$ and $\delta(\beta, l(F_i))  = 0 $ otherwise.

\begin{figure*}[htbp]
    \centering 
\captionsetup[subfigure]{labelformat=empty}
\begin{subfigure}{0.2\textwidth}
  \includegraphics[width=\linewidth]{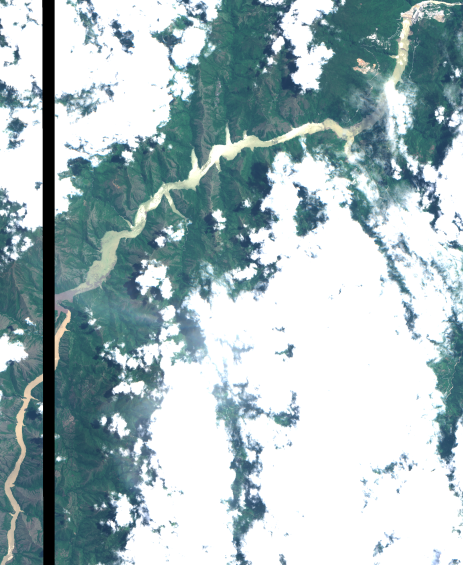}
\end{subfigure}\hfil 
\begin{subfigure}{0.2\textwidth}
  \includegraphics[width=\linewidth]{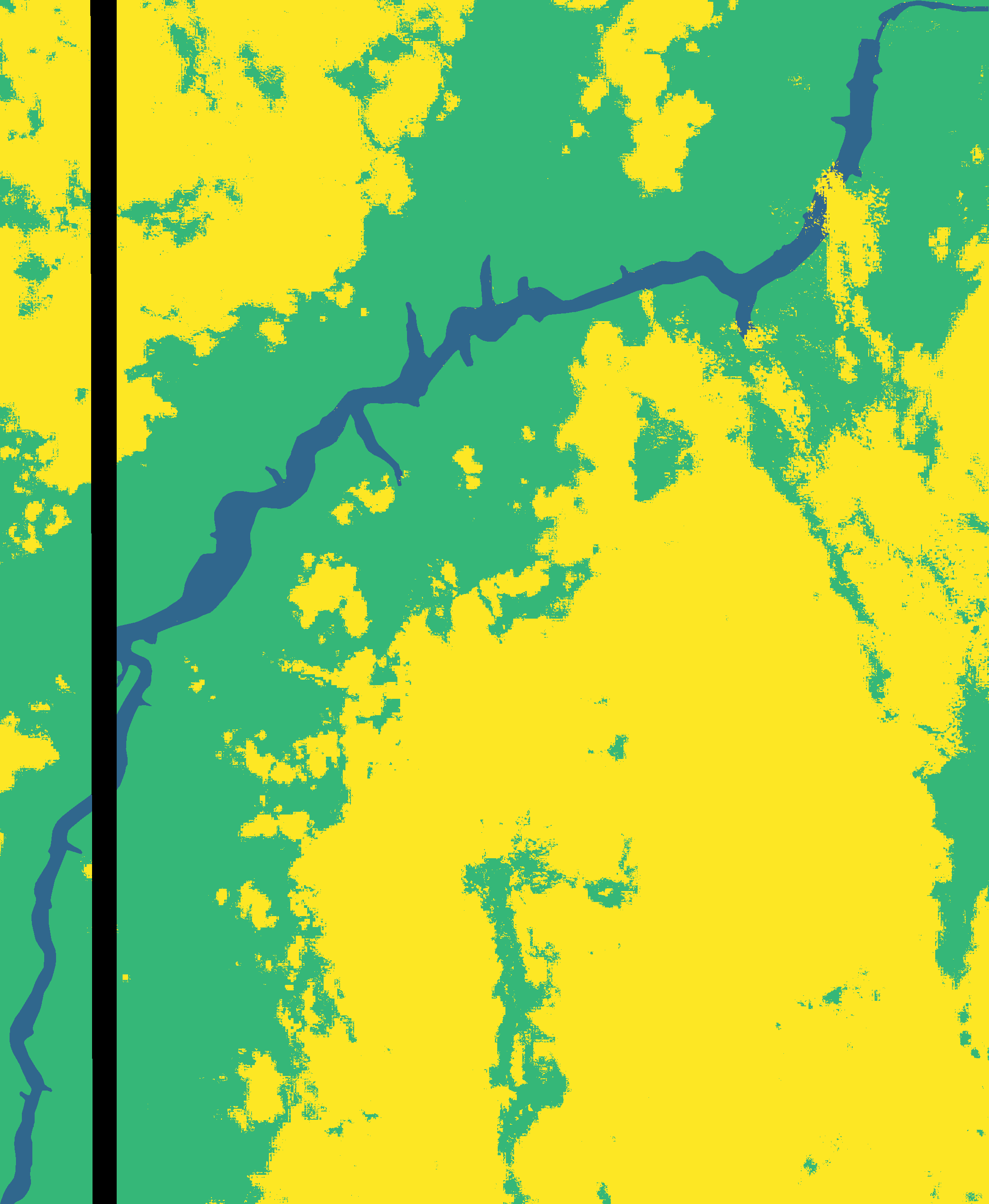}
\end{subfigure}\hfil 
\begin{subfigure}{0.2\textwidth}
  \includegraphics[width=\linewidth]{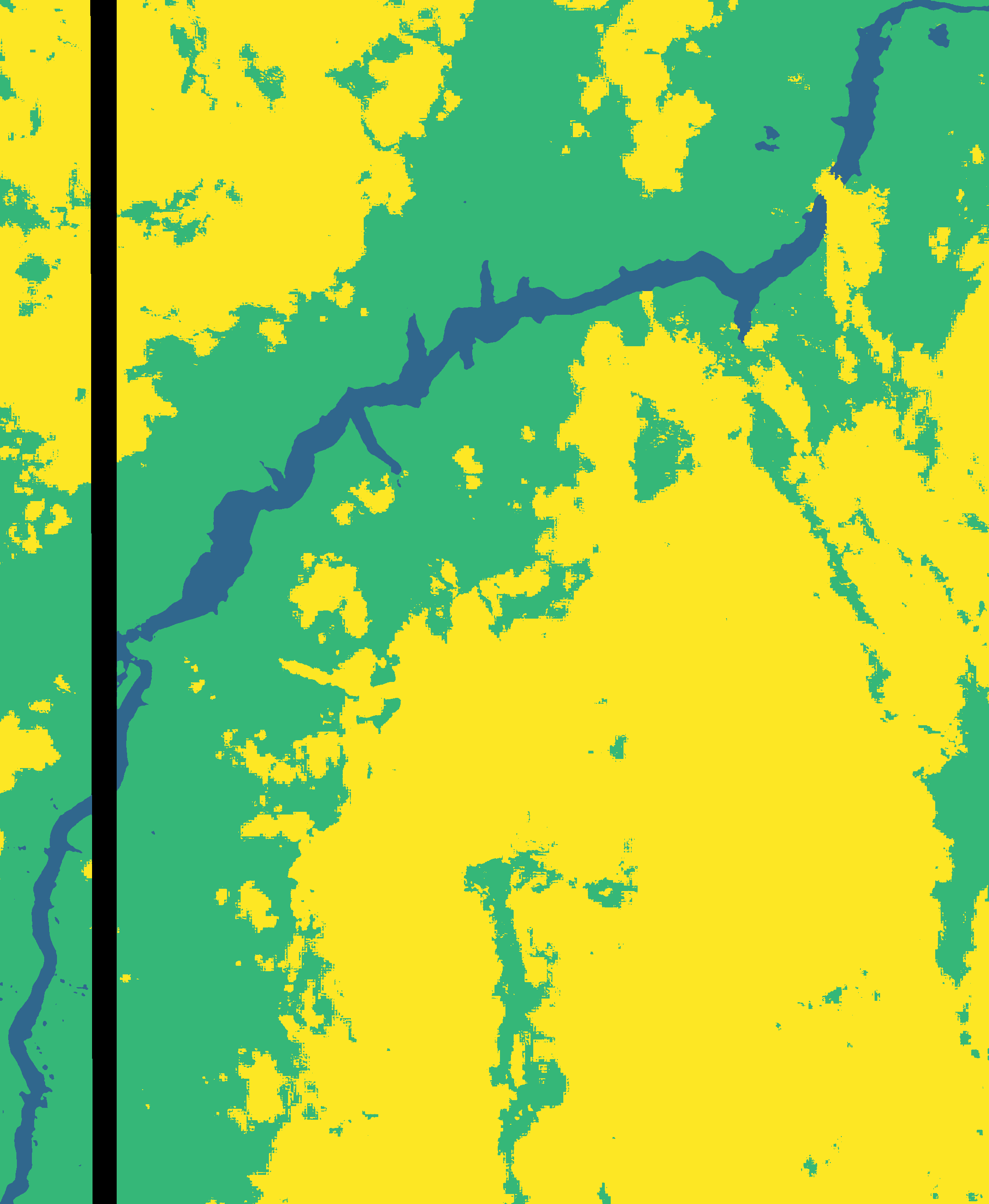}
\end{subfigure}\hfil 
\begin{subfigure}{0.2\textwidth}
   \includegraphics[width=\linewidth]{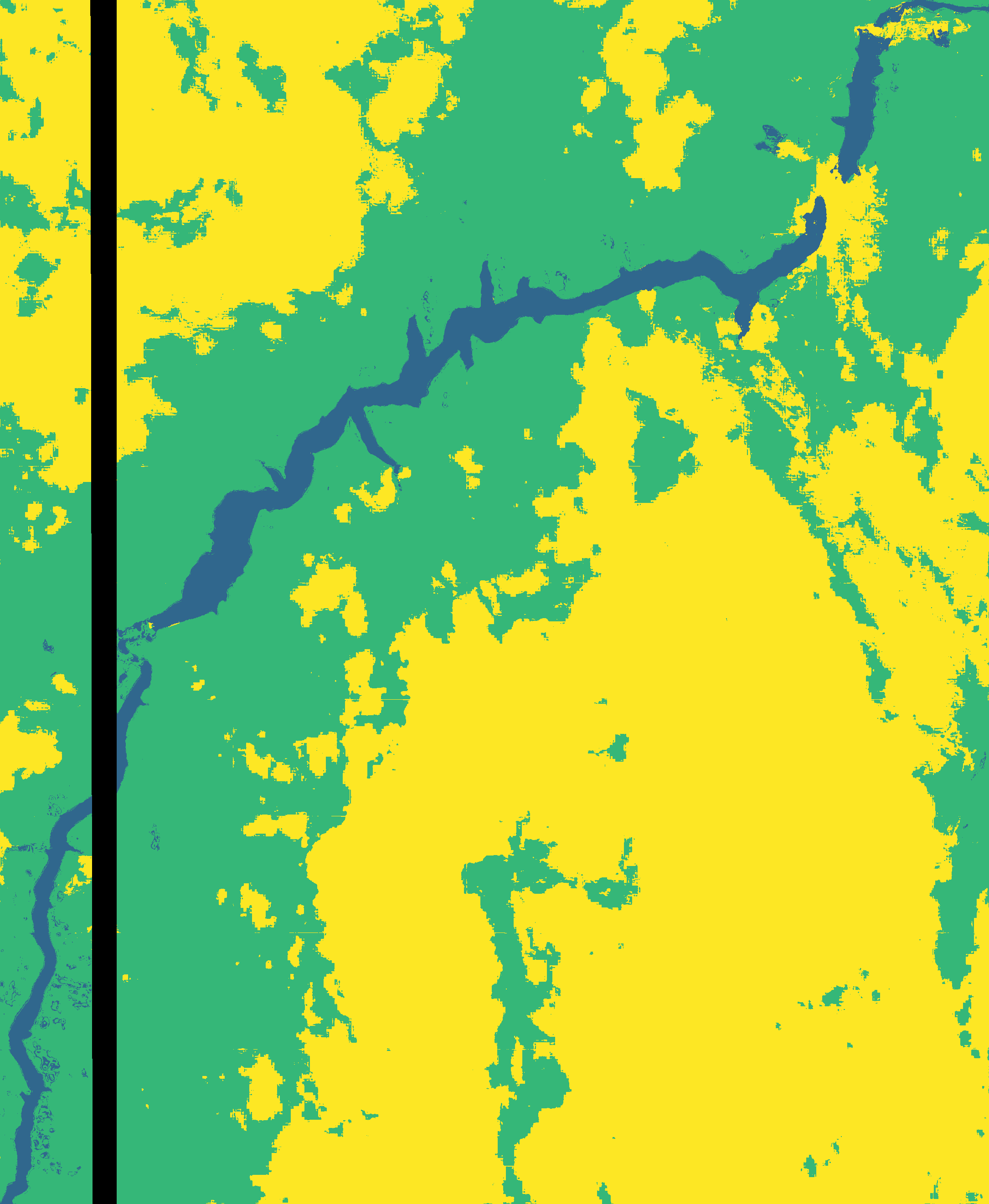}
\end{subfigure}

\medskip
\begin{subfigure}{0.2\textwidth}
  \includegraphics[width=\linewidth]{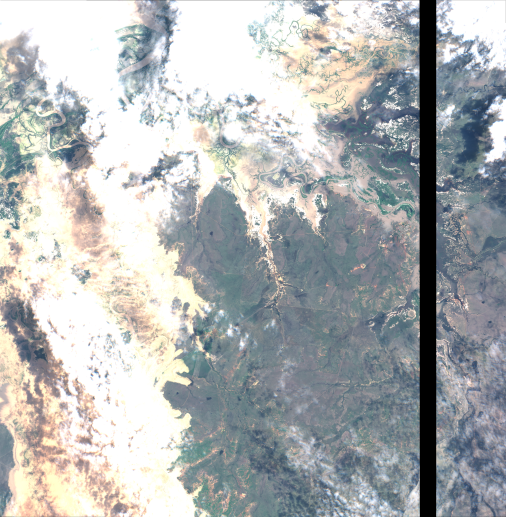}
\end{subfigure}\hfil 
\begin{subfigure}{0.2\textwidth}
  \includegraphics[width=\linewidth]{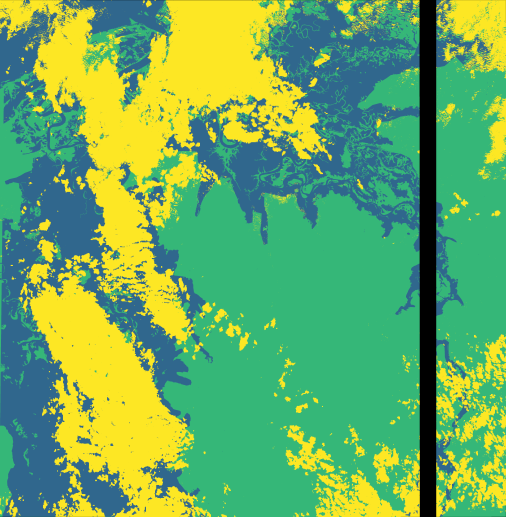}
\end{subfigure}\hfil 
\begin{subfigure}{0.2\textwidth}
  \includegraphics[width=\linewidth]{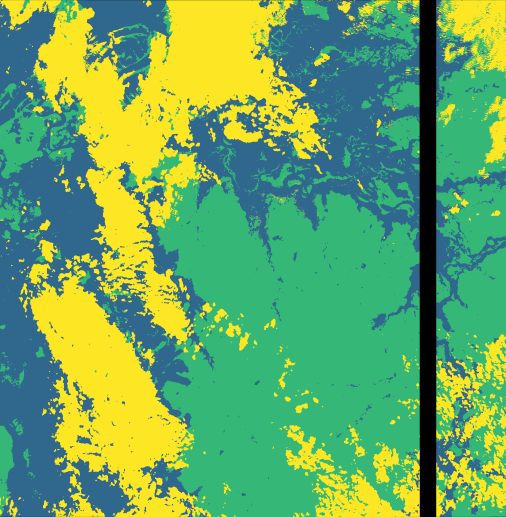}
\end{subfigure}\hfil 
\begin{subfigure}{0.2\textwidth}
  \includegraphics[width=\linewidth]{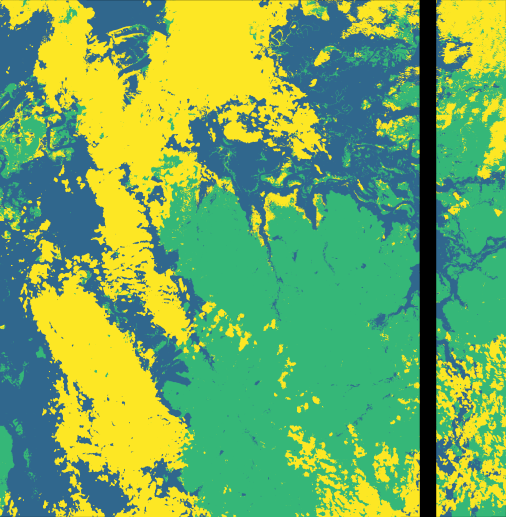}
\end{subfigure}

\medskip
\begin{subfigure}{0.2\textwidth}
  \includegraphics[width=\linewidth]{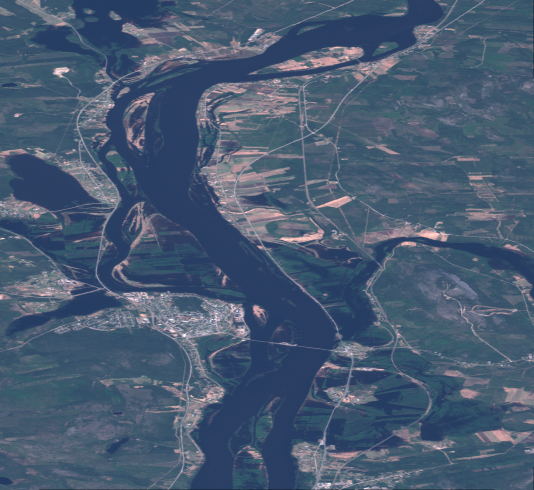}
\end{subfigure}\hfil 
\begin{subfigure}{0.2\textwidth}
  \includegraphics[width=\linewidth]{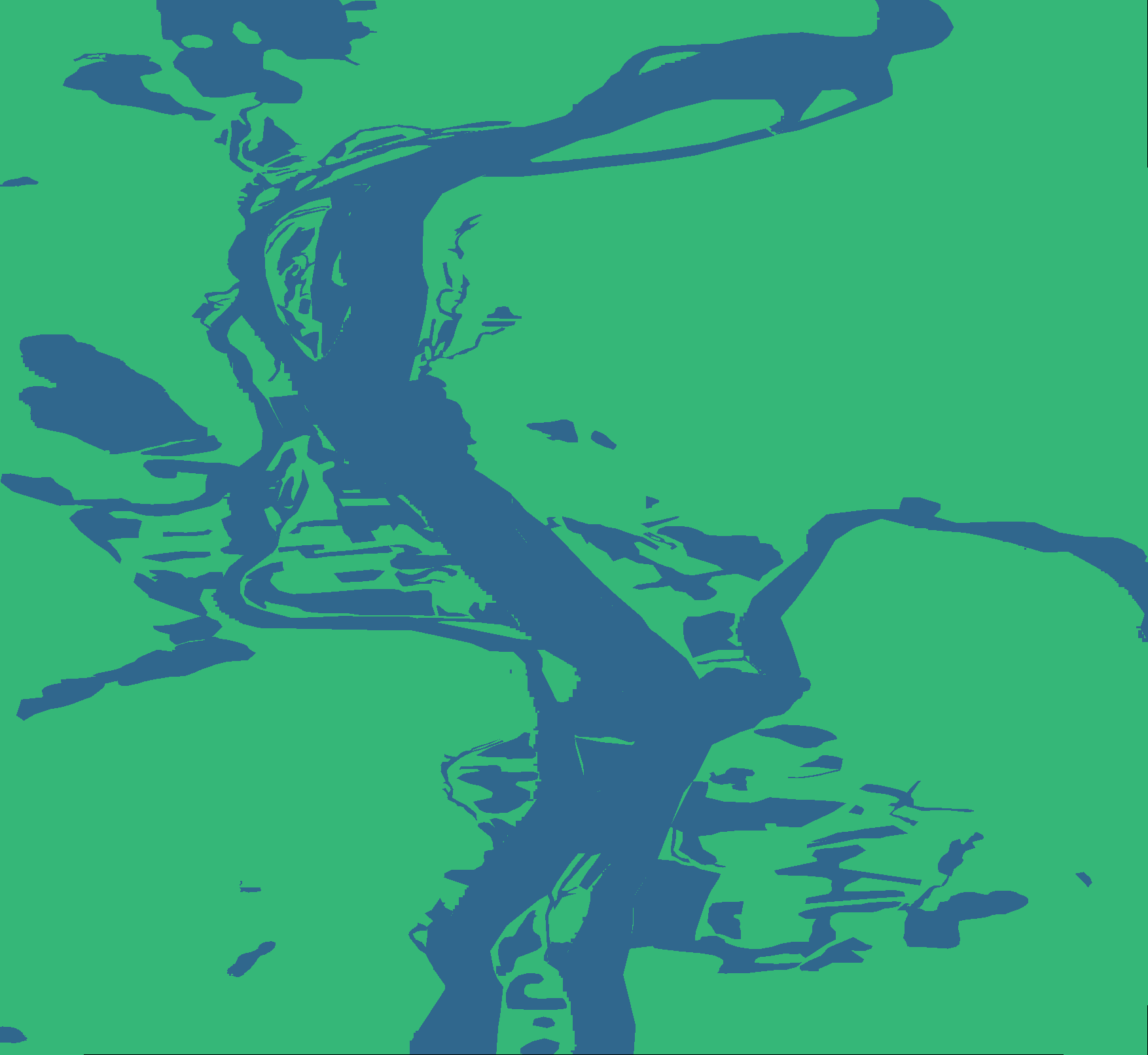}
\end{subfigure}\hfil 
\begin{subfigure}{0.2\textwidth}
  \includegraphics[width=\linewidth]{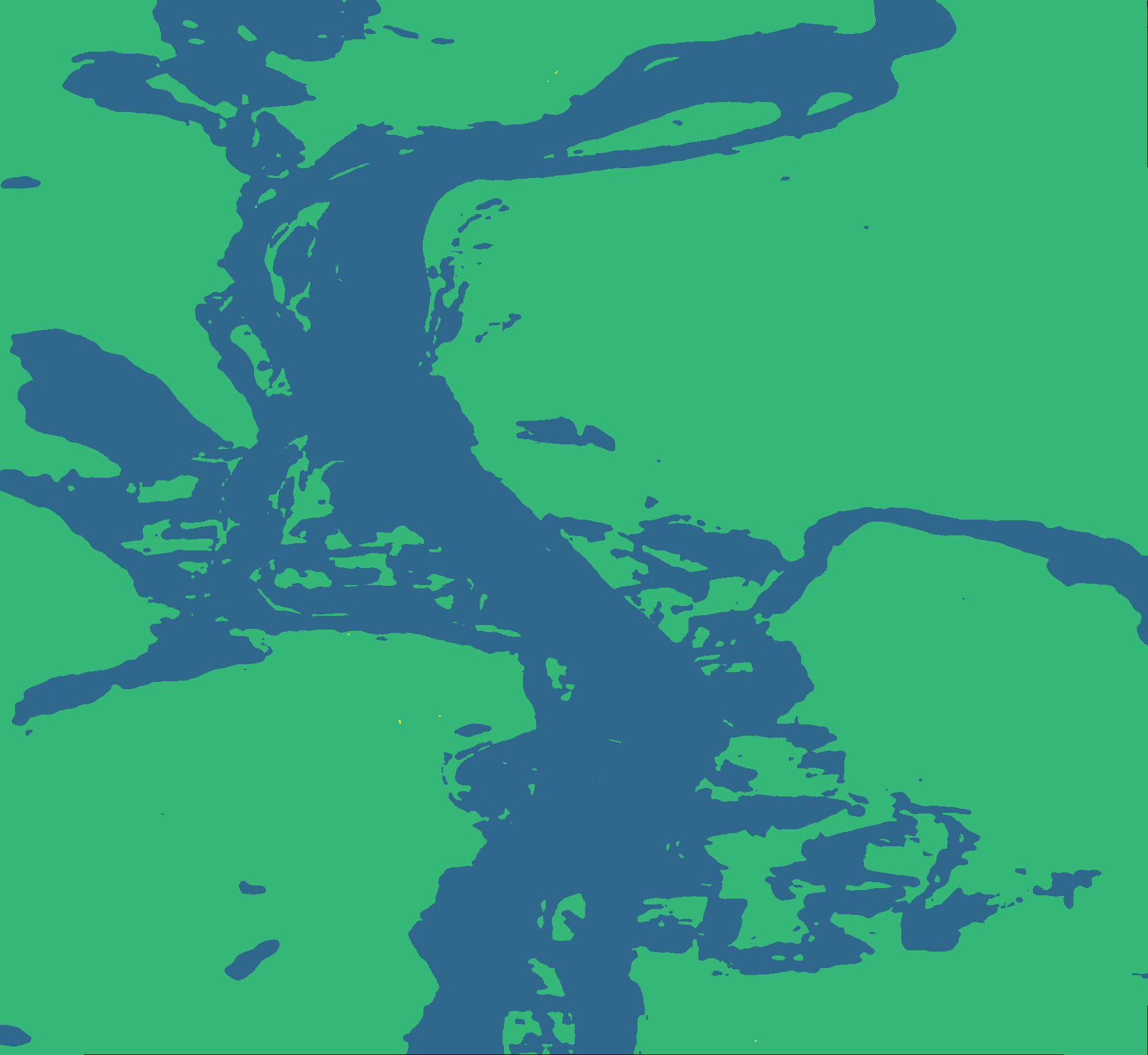}
\end{subfigure}\hfil 
\begin{subfigure}{0.2\textwidth}
  \includegraphics[width=\linewidth]{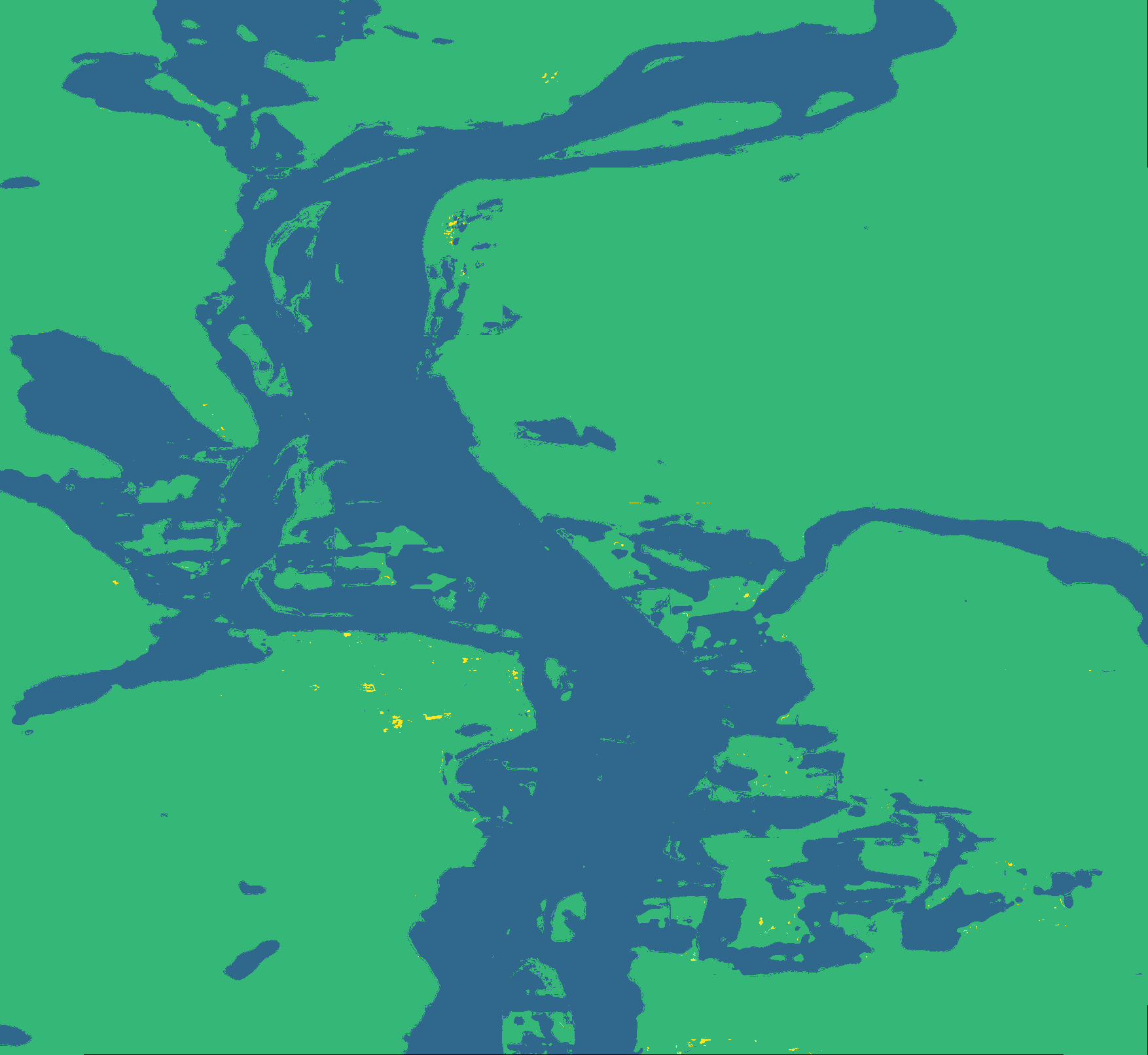}  
\end{subfigure}

\medskip
\begin{subfigure}{0.2\textwidth}
  \includegraphics[width=\linewidth]{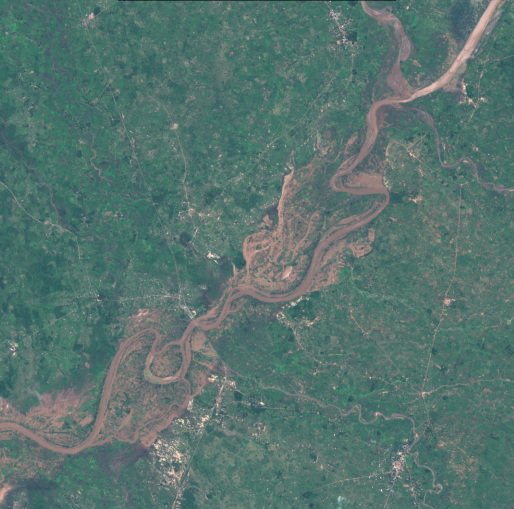}
\end{subfigure}\hfil 
\begin{subfigure}{0.2\textwidth}
  \includegraphics[width=\linewidth]{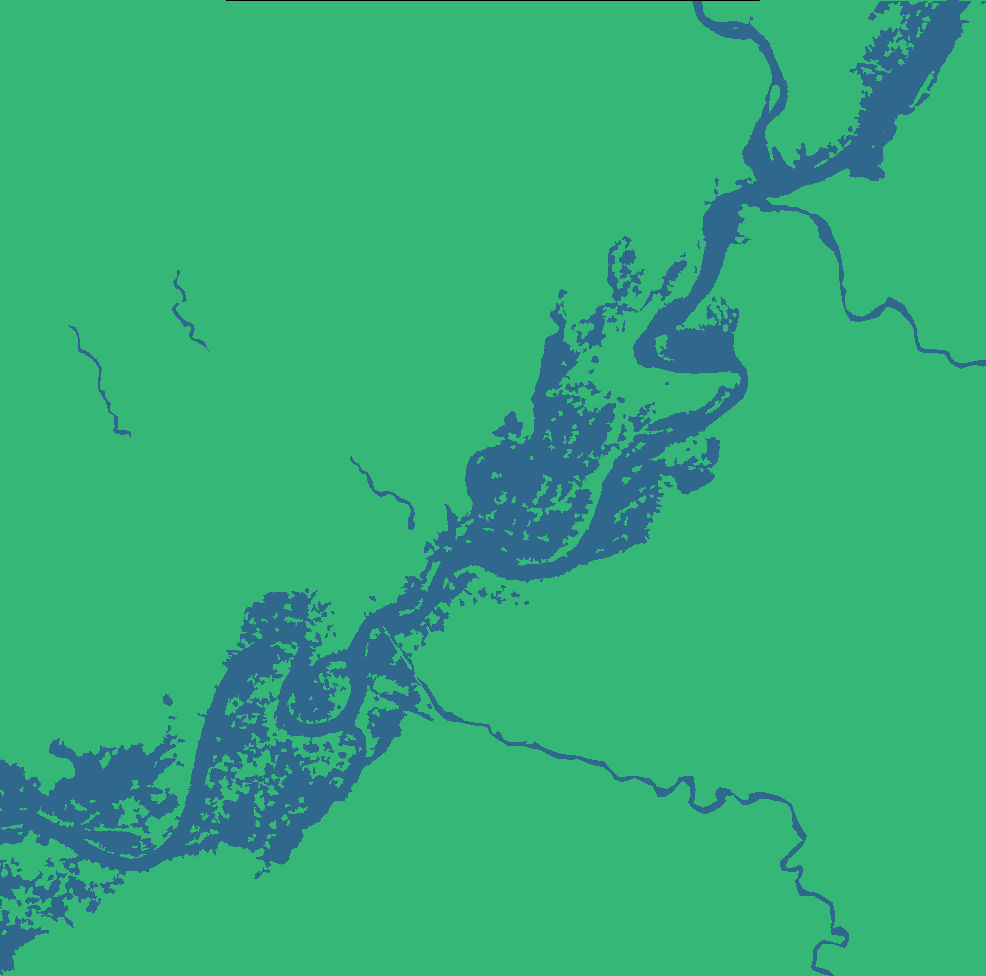}
\end{subfigure}\hfil 
\begin{subfigure}{0.2\textwidth}
  \includegraphics[width=\linewidth]{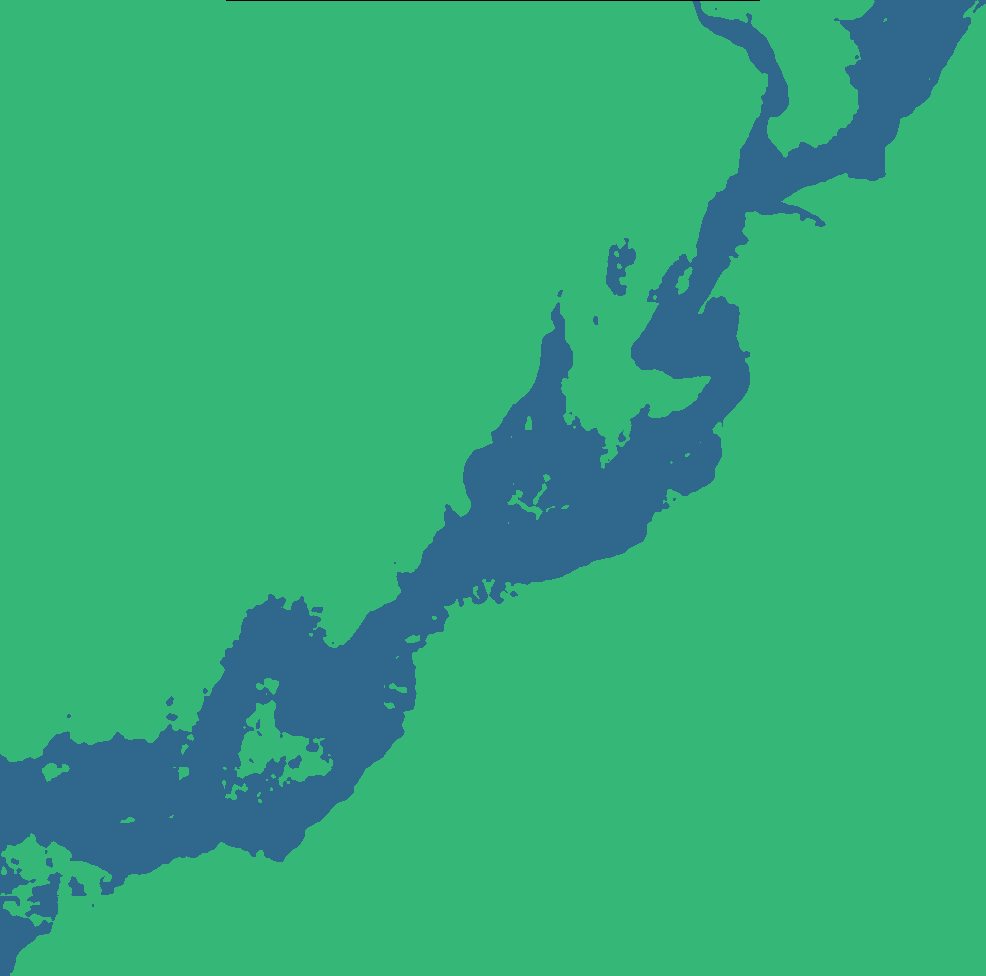}
\end{subfigure}\hfil 
\begin{subfigure}{0.2\textwidth}
  \includegraphics[width=\linewidth]{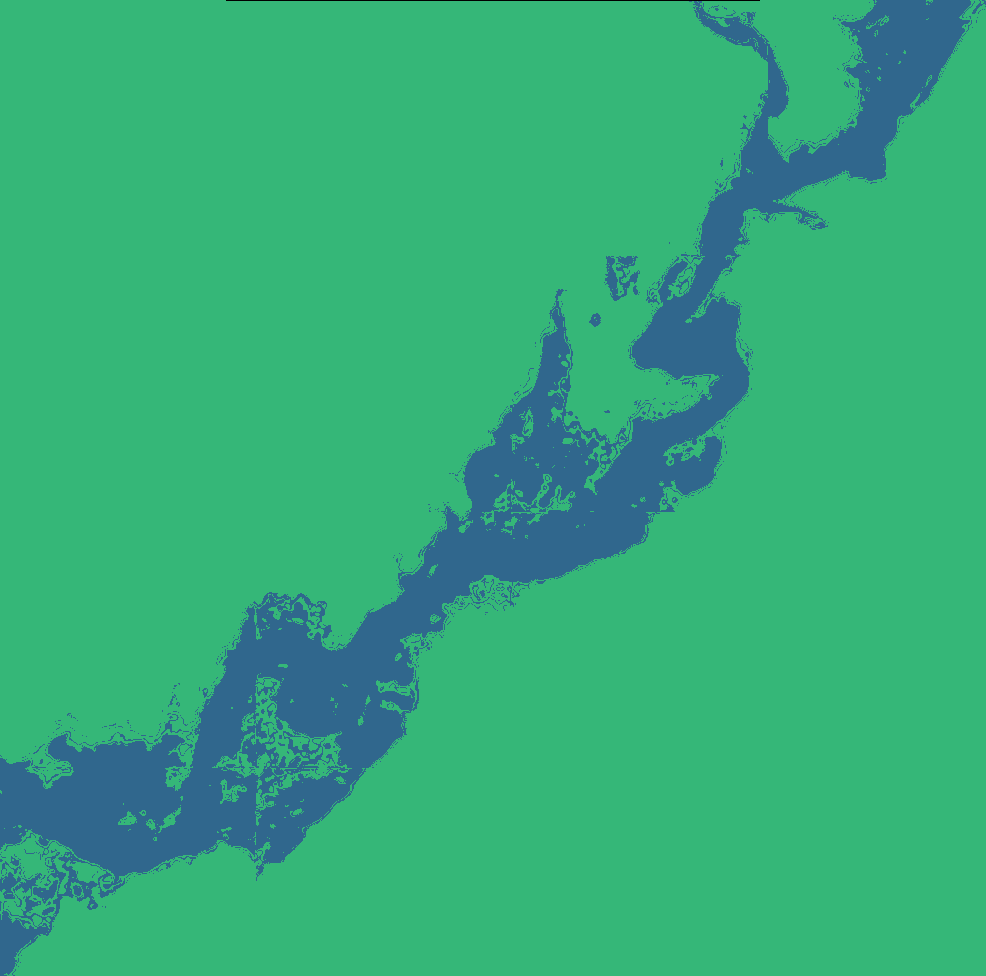}  
\end{subfigure}

\medskip
\begin{subfigure}{0.2\textwidth}
  \includegraphics[width=\linewidth]{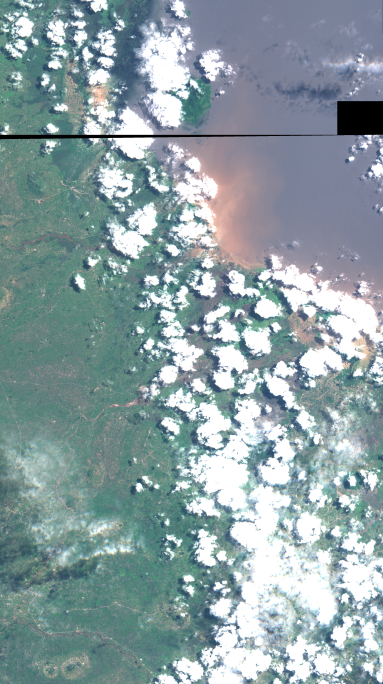}
  \caption{RGB}
\end{subfigure}\hfil 
\begin{subfigure}{0.2\textwidth}
  \includegraphics[width=\linewidth]{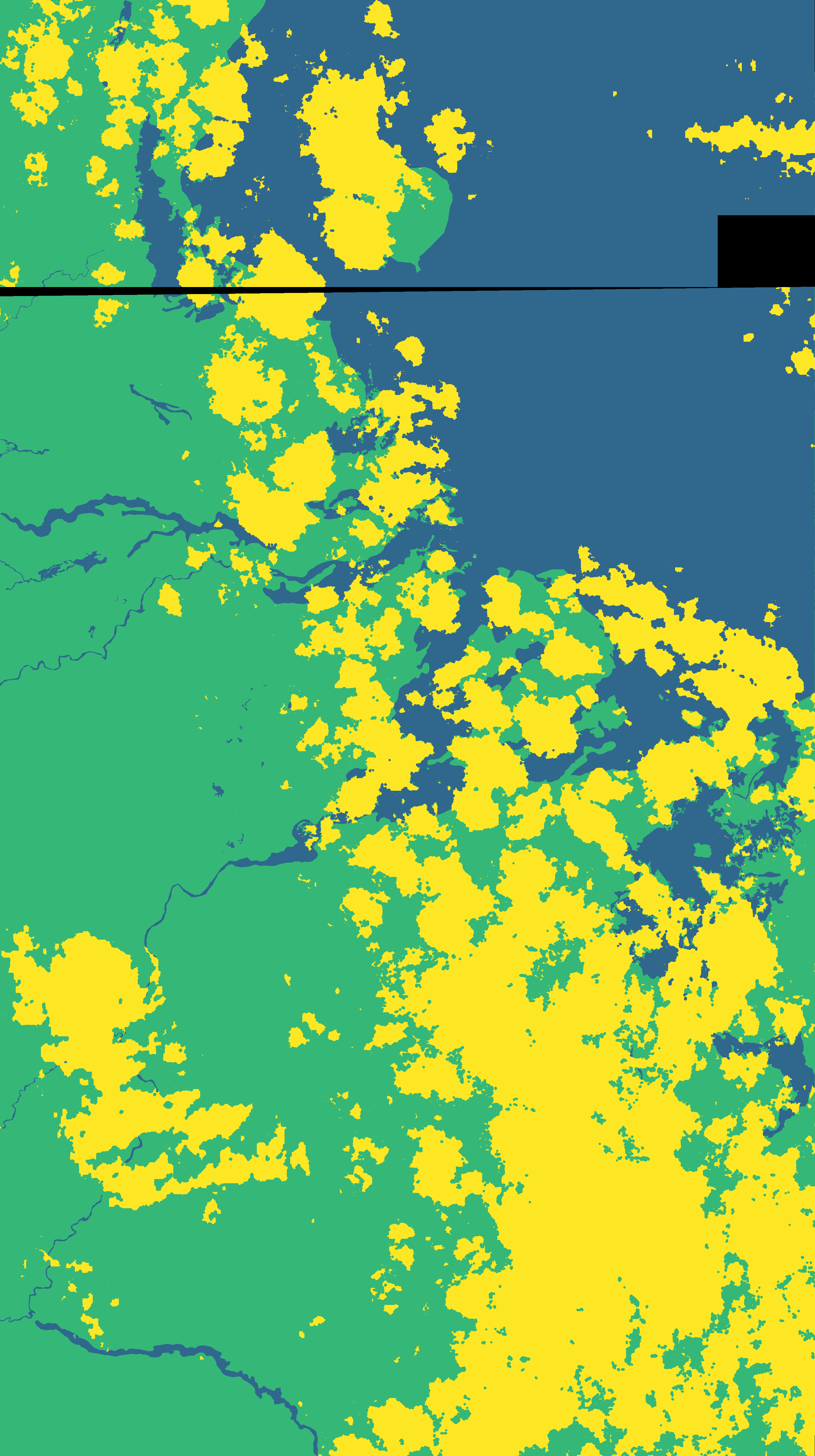}
  \caption{Labels}
\end{subfigure}\hfil 
\begin{subfigure}{0.2\textwidth}
  \includegraphics[width=\linewidth]{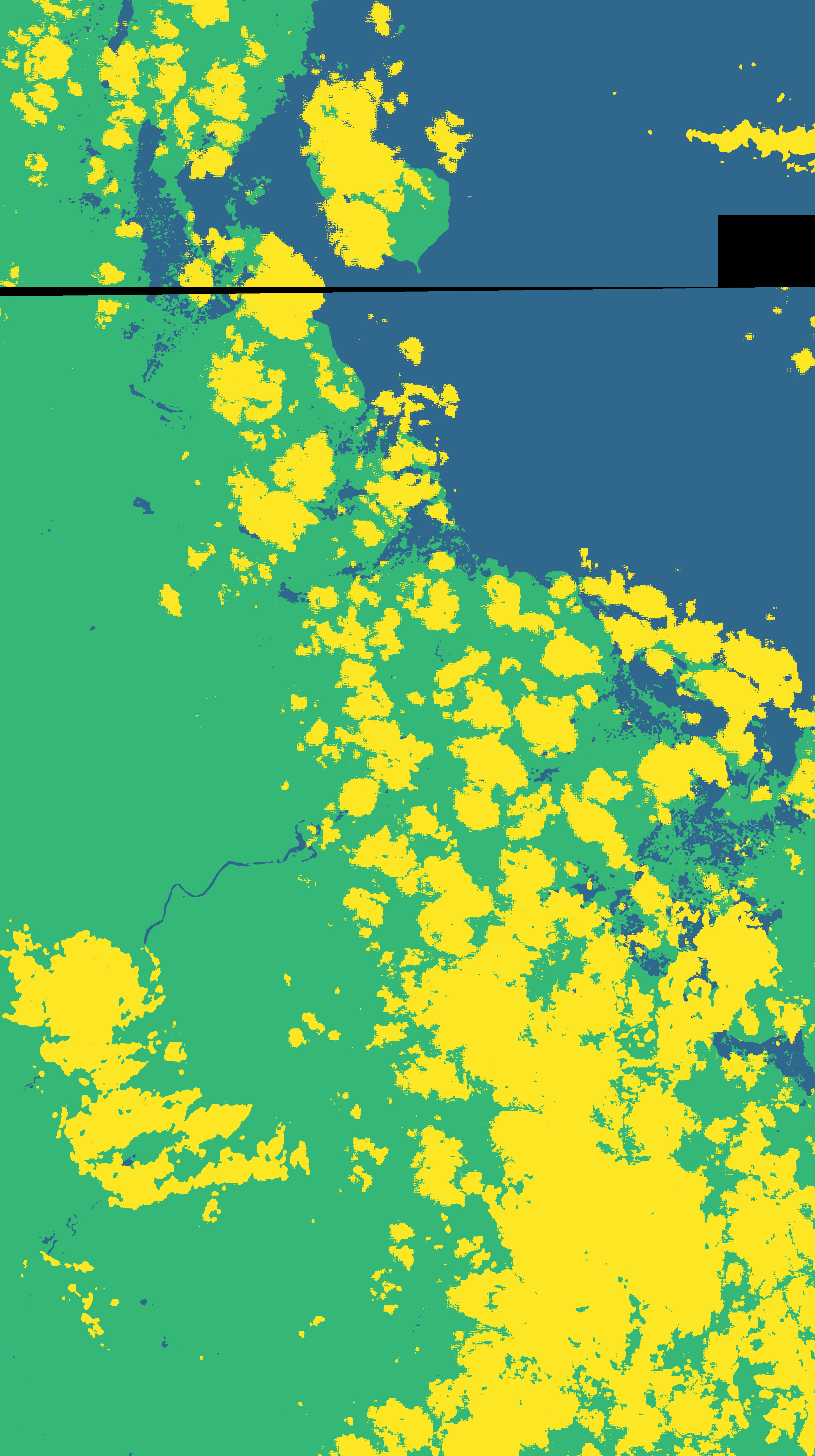}
  \caption{U-Net}
\end{subfigure}\hfil 
\begin{subfigure}{0.2\textwidth}
  \includegraphics[width=\linewidth]{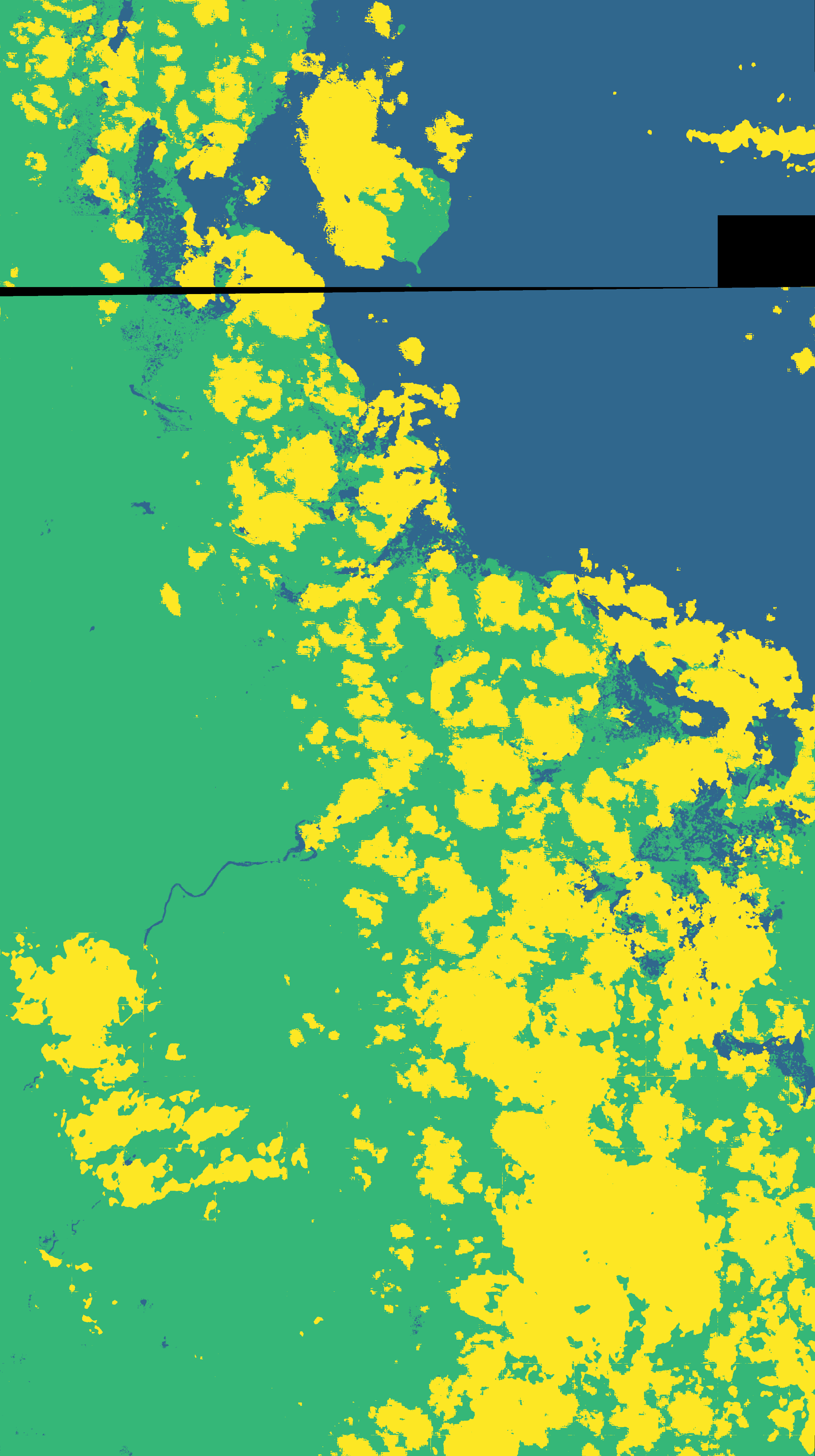}
  \caption{Ours}
\end{subfigure}
\caption{Comparison of segmentation results. The meaning of the colours is shown below: green - land, yellow - cloud, blue - water.}
\label{results}
\end{figure*}

\section{Experiment setup}
\subsection{Worldfloods Dataset}

In this paper to test the newly proposed IDSS method the rich and recent Worldfloods dataset \cite{mateo2021towards} is used. It describes 119 real flood events and has 424 flood maps. All flood maps are based on Sentinel-2 images containing 13 bands which form the raw features vector. The description of each channel is shown in Table \ref{Table1} \cite{drusch2012sentinel} \cite{phiri2020sentinel} and a statistical analysis of the data is shown in Table \ref{Table2} \cite{mateo2021towards}. The bands with a resolution larger than 10m were downsampled to 10m using nearest neighbours interpolation. From Table \ref{Table2} it can be observed that the percentage of could pixels in the training data set is much higher than that in the validation and test data sets. Such an anomaly also takes place for the  land pixel from the validation data set and the flood water pixels from the test data set. The Worldfloods data set is highly imbalanced (with a very small proportion of water pixels), which is a serious challenge for the classification task.

\begin{table*}[htb]
\caption{Statistics of the Worldfloods dataset}
\begin{center}
\begin{tabular}{|c|c|c|c|c|c|c|c|}
\hline
\textbf{Dataset} & \textbf{\textit{Flood events}} & \textbf{\textit{Flood maps}}&\multicolumn{2}{|c|}{\textbf{\textit{Water pixels (\%)}}}& \textbf{\textit{Land pixels}}& \textbf{\textit{Cloud pixels}}& \textbf{\textit{Invalid pixels}} \\
\hline
 &  &   & Flood & Permanent & (\%) & (\%)& (\%)  \\
\hline
Training & 108 & 407 & 1.45 & 1.25 & 43.24 & 50.25&3.81  \\
\hline
Validation & 6 & 6 & 3.14 & 5.19 & 76.72 & 13.27 & 1.68 \\
\hline
Test & 5 & 11 & 20.23 & 1.16 & 59.05 & 16.21 & 3.34 \\
\hline
\end{tabular}
\end{center}\label{Table2}
\end{table*}

\subsection{Training}
The implementation of the proposed model is based on the ML4Floods pipeline \cite{mateo2021towards} and scikit-learn \cite{pedregosa2011scikit}. The U-Net has been pre-trained on the Worldfloods dataset and used for extraction of the latent features. The Sentinel-2 images from the training data set were cropped to size 256 $\times$ 256 for feature extraction via U-Net. Different from the convolutional neural network, the proposed IDSS does not require a validation data set to check the model performance. In this paper the 6 validation Sentinel-2 images were used as a training data set which also demonstrates the ability of the proposed method to work with a smaller labelled training data sets.

\subsection{Testing}
The test images are cut into 256 $\times$ 256 for feature extraction. It should be noted that Worldfloods images vary in size significantly. Most of the images cannot be cut entirely to the size of 256 $\times$ 256. The patches that are with a smaller size than 256 $\times$ 256 are discarded in training, but this can not be done during testing. In this paper padding is performed on the test images so that they can be perfectly cut into 256$ \times$ 256 patches. After that, the prediction masks are stuck together, the padding part is simply being cut off. All experiments in this paper were conducted on the High-End Computing Cluster with Intel(R) Xeon(R) Gold 6248 CPU @ 2.50 GHz processor.

\subsection{Performance Evaluation}

The following metrics were used to evaluate the proposed algorithm:

Intersection over Union (IoU):
\begin{equation}
IoU = \dfrac{TP}{FP+TP+FN}
\end{equation}

Recall:
\begin{equation}
Recall = \dfrac{TP}{TP+FN}
\end{equation}

Here, TP, FP and FN represent True Positive, False Positive and False Negative, respectively.
IoU is the widely used measure of success for this problem because it maximizes the TP while minimizing FP as well as TP and FN. Recall is considered to be the more important measure than Precision since missing flood areas are more dangerous and risky than over-predicting flood areas.

\section{Results}

The results of the proposed IDSS method are presented in Table \ref{Table3} and summarized in this section. 

\begin{table}[h]
\caption{Comparison of IoU and recall results}
\begin{center}
\resizebox{\columnwidth}{!}{  
\begin{tabular}{|c|c|c|c|}
\hline
\multirow{2}*{Model} & {\textit{IoU} total water}& {\textit{Recall} total water} & \# Parameters\\

 & \%  & \% & (x1000) \\ \hline
$IDSS^1$ (ours) & \textbf{\underline{73.10}}& 93.35 & 96\\ 
$IDSS^2$ (ours) & 70.03 & \textbf{\underline{96.11}} & 96\\
xDNN \cite{angelov2020towards} & 71.50 & 90.27 & 20\\
U-Net \cite{mateo2021towards} & 72.42& 95.42 & 7790\\
SCNN \cite{mateo2021towards} & 71.12 & 94.09 & 260\\
$NDWI^1$ \cite{mcfeeters1996use} & 65.12 & 95.75 & - \\
Linear \cite{mateo2021towards} & 64.87 & 95.55 & -\\
$NDWI^2$ \cite{mcfeeters1996use} & 39.99& 44.84 & -\\
\hline
\end{tabular}
}
\label{Table3}
\end{center}
\end{table}

Table \ref{Table3} shows the metric comparison between the proposed algorithms and a number of state-of-the-art methods, including threshold-based and Deep Neural Network algorithms. It can be observed that the proposed algorithm achieves very competitive results going beyond the state-of-the-art. While trained only on six images, the proposed method $IDSS^1$ has 0.68\% IoU total water improvement compared to U-Net. At the same time, the proposed \textit{IDSS} method also offers linguistic form of explainability (Fig. \ref{Linguistic}) and a clear network architecture (Fig. \ref{TrainingF} and Fig. \ref{ValidationF}) and is using in orders of magnitude less model parameters. $IDSS^2$ trained on 206 training images has 0.36\% Recall total water improvement  compared to the second highest method $NDWI^1$. The threshold used for $NDWI^1$ was $-0.22$ and for $NDWI^2$ it was $0$.

xDNN is an interesting recently proposed explainable form of Deep Neural Network method which in this paper was trained on the raw Sentinel-2 images without feature extraction and, again, only trained on six images. In this paper before applying this method the Worldfloods model was used to remove the clouds. the result using xDNN after cloud removal is also competitive with the IoU total water of 71.50\%. More importantly, however, the xDNN algorithm provides direct interpretability through the linguistic \textit{IF ... THEN} rules and the decision making can be understood and analyzed by humans, which can play an important role in practical applications. 

A visual comparison of the segmentation results is shown in Fig. \ref{results}. It can be appreciated visually that the segmentation results obtained by the proposed method are very similar to those obtained by the U-Net model and to the reference labels. U-Net model tends to over-predict water areas if compare the images in the fourth row.

\section{Conclusion}

In this paper, an interpretable deep semantic segmentation (IDSS) method is proposed for Earth Observation. It provides human interpretable results while ensuring a performance on par or surpassing that of the mainstream and state-of-the-art convolutional neural network. The proposed IDSS is a prototype-based method. Prototypes are the local focal points that are most representative in terms of the features. Working in a latent feature space and a projection to the raw feature space allows a human interpretable linguistic \textit{IF...THEN} rules to be used to explain the decision making involved in determining the winning class label. Semantic rules are interpretable and easy for humans to understand allowing users could inspect and audit the decision making process. Worldfloods dataset was used, and the results show that IDSS outperforms the other methods in terms of IoU total water and Recall total water requiring in orders of magnitude less parameters and training data. The future work will focus on investigation of the application of the model to flood detection.

\bibliographystyle{IEEEtran}
\bibliography{conference_101719}
\end{document}